\pdfoutput=1
\documentclass[a4paper,11pt]{article}

\usepackage[margin=1in]{geometry}
\usepackage[T1]{fontenc}
\usepackage[utf8]{inputenc}
\usepackage{ebgaramond}
\usepackage{amsmath}
\usepackage{amssymb}
\usepackage{microtype}
\usepackage{graphicx}
\usepackage{caption}
\usepackage{subcaption}
\usepackage{booktabs}
\usepackage{placeins}
\usepackage{url}
\usepackage{hyperref}

\usepackage{titlesec}
\titleformat{\section}
  {\fontsize{16}{19}\selectfont\bfseries}
  {}{0pt}{}
\titlespacing*{\section}{0pt}{8pt}{4pt}

\makeatletter
\newcommand{\affiliations}[1]{\def\@affiliations{#1}}
\affiliations{}
\renewcommand{\maketitle}{%
  \begin{center}
    {\fontsize{26}{30}\selectfont \@title \par}
    \vspace{4pt}
    {\fontsize{10}{12}\selectfont \@author \par}
    \vspace{4pt}
    {\fontsize{9}{11}\selectfont \@affiliations \par}
  \end{center}
  \vspace{8pt}
}
\makeatother

\begin{document}

\title{Benchmarking Label-Revealed Online Updates for EEG BCI Decoding}

\author{Bogdan Kozyrskiy$^{1}$, Artem Grachev$^{2}$, Abraham I. Camelo Guerrero$^{3}$}

\affiliations{%
$^{1}$Independent Researcher, France\\
$^{2}$Independent Researcher, Germany\\
$^{3}$School of Information Technology, York University, Toronto, Canada%
}

\maketitle

\section{Abstract}
Electroencephalography (EEG) signals drift over time, which can cause static brain--computer interface (BCI) models to degrade in practice. We present a benchmark for online adaptation and compare two widely used pipeline families, Common Spatial Patterns (CSP) and Riemannian covariance-based methods, under time-ordered prequential (test-then-train) evaluation. We examine (i) which pipelines benefit most from label-revealed updates, (ii) whether controlled forgetting of older data improves robustness, and (iii) how a minimal-calibration cold start compares with starting from a pretrained model. Across four datasets (three motor-imagery datasets and one movement-decoding dataset), label-revealed online updates improve 13 of 14 model/dataset pairs on the two largest streams, with relative accuracy gains of up to about 18\% over a frozen model. A Shapley-based data-valuation analysis over temporal blocks assigns the largest mean value to the most recent block in each of the three analyzed datasets, while older blocks retain positive value.

\textbf{Keywords:} Online adaptation, Continual learning,
Prequential evaluation, Shapley-based data valuation

\section{Introduction}
A Brain-Computer Interface (BCI) is a technology that provides direct communication between the human brain and external devices. The system works by acquiring brain signals via electroencephalography (EEG) that are subsequently analyzed for the detection of specific mental activities \cite{Varbu2022}. BCIs can assist people with motor impairments \cite{Guo2022} and enable control of external devices \cite{Padfield2022}. Improving BCI decoding methods can therefore support both practical assistive technologies and a better understanding of brain function.

Motor imagery (MI), the cognitive process of imagining performing a movement, has been extensively studied in the machine learning (ML) community. In BCI research, ML pipelines are commonly used for feature extraction and classification \cite{Hou2022}. Traditionally,
Common Spatial Patterns (CSP) and Riemannian covariance-based methods have been used for feature extraction, followed by a classifier. BCI models are often evaluated offline, but real EEG data arrive sequentially and can drift over time. As a result, the signal patterns associated with a given task may change \cite{Vavoulis2023}. Online adaptation methods address this by updating the model as new labeled trials become available.

This work proposes a benchmark for online adaptation in BCI. Performance is measured using prequential evaluation (test-then-train), which captures how the model adapts over time. We report cumulative prequential accuracy, averaged across subjects, as the primary metric to enable consistent comparisons across datasets with different numbers of classes.

\textbf{Contributions.} This paper contributes: (i) a prequential, label-revealed benchmark protocol with four evaluation/update regimes (frozen/offline, incremental/expanding-window, rolling-window, and cold start); (ii) an empirical comparison of seven CSP- and covariance-based pipelines across four EEG decoding datasets; (iii) a per-step prequential analysis, using cumulative ($C_k$) and trailing-window ($A_k$) accuracy, that contrasts frozen with online-refit decoders under paired significance tests; and (iv) a Shapley-based temporal data-valuation analysis that quantifies the value of recent versus older trials and replicates it across datasets.

\section*{Experiments and methods}
\subsection*{Pipelines}
We reuse a subset of the reference pipelines from the MOABB benchmark study~\cite{chevallier2024moabb} and evaluate them in a prequential continual-learning protocol. Each pipeline is a composition of (i) fixed signal preprocessing, (ii) a feature-extraction block, and (iii) a classifier that is updated online after the true label is revealed.

\textbf{Preprocessing (common).}
EEG is segmented into epochs and band-pass filtered (4--38 Hz) to emphasize task-relevant rhythms and standardize inputs across methods~\cite{chevallier2024moabb}. This preprocessing is kept fixed throughout the stream: epoching, filtering, and channel selection are not adapted online. For all
datasets, we additionally discard the first second after stimulus onset to mitigate onset-related artifacts and reduce dimensionality. The epoch length is otherwise left at the dataset default (i.e., we use each dataset’s standard evaluation window).

\textbf{Riemannian covariance and tangent-space features (TS).}
Riemannian pipelines represent each trial with a symmetric positive definite (SPD) covariance matrix and exploit the geometry of the SPD manifold for robust decoding~\cite{chevallier2024moabb, Congedo2017RiemannianPrimer}. To enable standard vector-space classifiers, covariance matrices are mapped to the tangent space at a reference point (typically the Riemannian mean) and vectorized into Euclidean features (TS)~\cite{chevallier2024moabb,pyRiemann2023}. We evaluate TS with linear models (logistic regression, Elastic Net) and SVMs (linear and RBF), following common practice in the literature and the MOABB reference pipelines~\cite{chevallier2024moabb}.

\textbf{Augmented Covariance Matrix (ACM).}
ACM extends the covariance representation by incorporating time structure through a delay-embedding / autoregressive view of the signal, yielding a larger SPD matrix that captures spatio-temporal dependencies~\cite{Carrara2023ACM}. In practice, ACM is controlled by an embedding order and lag (grid-searched in the reference implementation) and is then processed with the same TS block and classifier as above (ACM+TS+SVM)~\cite{chevallier2024moabb}.

\textbf{Spatial filtering features: CSP and FBCSP.}
CSP learns supervised spatial filters that maximize variance differences between classes; log-variance of the filtered components forms the feature vector used by a downstream classifier (CSP+LDA, CSP+SVM)~\cite{chevallier2024moabb,Blankertz2008CSP}. FBCSP improves robustness to subject-specific frequency variability by applying CSP independently within a filter bank and concatenating/selecting band-wise features before classification (FBCSP+SVM)~\cite{chevallier2024moabb,Ang2008FBCSP}.

\textbf{Classifiers and online updates.}
We use the same decision heads as the reference pipelines -- LDA or SVM for CSP features (SVM for FBCSP), and LR/Elastic Net/SVM for TS features~\cite{chevallier2024moabb}. In the continual-learning setting, feature extraction is recomputed per incoming trial. The classifier and feature extractor are updated using the currently available labeled data (e.g., full retraining on an expanding set or a sliding window), enabling adaptation under non-stationarity.

\subsection*{Datasets}
All datasets are accessed through the MOABB library. We evaluate each subject/session as a time-ordered stream of trials, consistent with our prequential (test-then-train) continual-learning protocol.

\textbf{BCI Competition IV-2a (BCIC 2a).}
This is a widely used four-class motor-imagery benchmark with 9 subjects recorded in 2 sessions on different days. Each trial is cue-based (left hand, right hand, feet, tongue) and EEG is recorded from 22 channels at 250\,Hz; each session contains multiple runs with repeated trials per class. We follow the dataset's standard epoching and labeling as provided by MOABB.

\textbf{Alex Motor Imagery (ALEXMI).}
AlexMI is a small, low-density motor-imagery dataset (8 subjects) recorded at 512\,Hz with 16 electrodes. Trials are 3\,s long and cover three classes (right-hand imagery, feet imagery, and rest), with 20 trials per class.

\textbf{OpenBMI Motor Imagery (Lee2019).}
This dataset contains left-vs-right hand grasp motor imagery from 54 participants, recorded with 62 channels at 1000\,Hz over 2 sessions. Each session includes labeled runs with balanced classes (100 trials per phase, in the original protocol). In our experiments we rely on labeled trials as exposed by MOABB, and preserve the within-run temporal order for streaming evaluation.

\textbf{High-Gamma / Movement Decoding (Schirrmeister 2017).}
This dataset comprises 14 subjects recorded with 128 EEG channels at 500\,Hz, with 4-second trials across four classes (left-hand finger tapping, right-hand finger tapping, toe clenching/both feet, and rest). In contrast to motor imagery paradigms, the task involves \emph{executed} movements cued by visual arrow stimuli, and was acquired in a setup optimized for high-frequency movement-related EEG components, making it a common benchmark for motor decoding. We use the trial structure and labels provided by MOABB and process trials sequentially in time.

\subsection*{Online Learning}

In realistic BCI operation, EEG trials for a given subject arrive sequentially, and model updates can only use data observed up to the current time. To emulate this deployment constraint on recorded data, we adopt prequential (interleaved test-then-train) evaluation: for each incoming trial, the model predicts the unseen label, the prediction is scored, and the model may then update using the newly revealed label. This trial-wise protocol differs from pseudo-online evaluation frameworks that approximate real-time processing by re-segmenting continuous recordings with overlapping sliding windows (e.g., \cite{carrara2024pseudo}). This protocol estimates how the model would have performed had it been deployed earlier and updated only with past trials. In this work, we refer to this setting as \emph{online (incremental) learning} and consider two common update schemes, together with a cold-start initialization variant.

\begin{figure}
    \centering
    \includegraphics[width=0.5\linewidth]{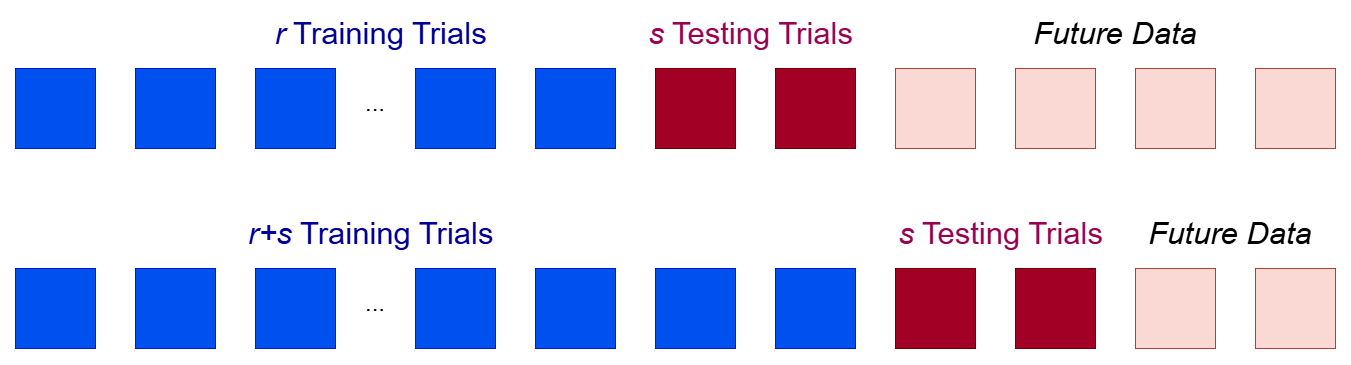}
    \caption{Expanding-window (walk-forward) evaluation scheme.}
    \label{fig:expanding}
\end{figure}

\textbf{Expanding window.}
The expanding-window (walk-forward) scheme is illustrated in Fig.~\ref{fig:expanding}. The model is trained on an initial set of $r$ trials, then predicts the labels of the next $s$ trials. After the corresponding true labels are revealed, the model is retrained using the enlarged training set containing the original $r$ trials plus the newly observed $s$ trials (i.e., $r \leftarrow r+s$), and the procedure repeats until the end of the stream. The standard incremental regime begins with the dataset’s training split, predicts one held-out trial ($s = 1$), reveals its label, appends that trial, and refits the pipeline before the next prediction. The frozen regime uses the same initial split but performs no refits. In the cold-start regime, the ordered splits are concatenated and the model starts from a short prefix: six trials for every pipeline except FBCSP, which uses 20 trials. Thus, the FBCSP cold-start column has a larger initial calibration set and should not be used for a cross-pipeline comparison of calibration efficiency.

\begin{figure}
    \centering
    \includegraphics[width=0.5\linewidth]{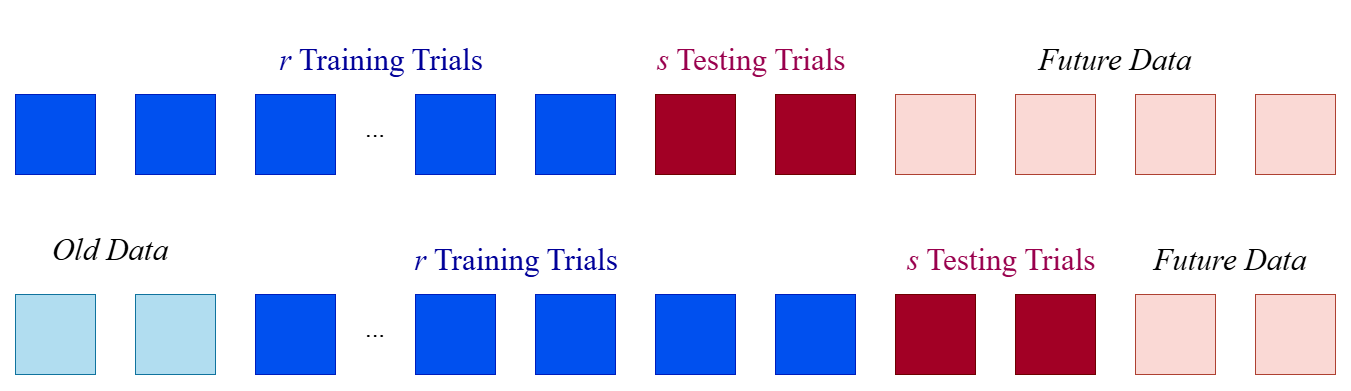}
    \caption{Rolling-window (sliding-window) evaluation scheme.}
    \label{fig:rolling}
\end{figure}

\textbf{Rolling window.}
The rolling-window scheme is depicted in Fig.~\ref{fig:rolling}. As in the expanding-window case, the model is trained on $r$ trials and predicts the next $s$ trials. However, the training set size is kept constant: after advancing by $s$ trials, the oldest $s$ trials are removed and the newest $s$ labeled trials are added, maintaining a window of $r$ trials at each iteration. This approach implements explicit forgetting of older data and can improve robustness under non-stationarity, when drift dominates the loss of sample size.

\textbf{Prequential accuracy.}
We summarize each stream with two prequential scores. The \emph{cumulative} accuracy after $k$ revealed labels,
\begin{equation}
  C_k = \frac{1}{k}\sum_{i=1}^{k}\mathbf{1}\{\hat{y}_i = y_i\},
  \label{eq:ck}
\end{equation}
measures the overall performance of the deployed system so far. The \emph{trailing-window} accuracy over the last $w$ predictions,
\begin{equation}
  A_k = \frac{1}{\min(k,w)}\sum_{i=\max(1,k-w+1)}^{k}\mathbf{1}\{\hat{y}_i = y_i\},
  \label{eq:ak}
\end{equation}
reflects the current performance of the system and is more sensitive to recent drift (we use $w=25$; for $k<w$ all available predictions are averaged). Unless stated otherwise, the overall accuracy reported below is the final cumulative score $C_K$ (with $K$ the number of streamed trials), averaged across subjects.

\FloatBarrier
\section*{Results}

\begin{table}[htbp]
\centering
\noindent
\begin{minipage}[t]{0.55\textwidth}
\centering
\captionof{table}{Overall Accuracy}
\label{tab:bcic}

\footnotesize
\setlength{\tabcolsep}{3pt}
\renewcommand{\arraystretch}{1}

\resizebox{\linewidth}{!}{%
\begin{tabular}{ l c c c c }
\hline
Models & Offline & Incremental & Window* & Cold Start \\
\hline
\multicolumn{5}{l}{\footnotesize \textbf{BCIC 2a}}\\
ACM+TS+SVM & 0.69 ± 0.11  & \textbf{0.77 ± 0.12} & 0.76 ± 0.13 & 0.71 ±0.13 \\
CSP+LDA    & 0.54 ± 0.13  & \textbf{0.61 ± 0.14} & 0.59 ± 0.13 & 0.57 ± 0.15 \\
CSP+SVM    & 0.60 ± 0.15  & \textbf{0.61 ± 0.14} & 0.60 ± 0.14 & 0.56 ± 0.15 \\
FBCSP      & 0.66 ± 0.15  & 0.69 ± 0.14 & \textbf{0.70 ± 0.15} & 0.64 ± 0.16 \\
TS+EL      & 0.59 ± 0.15  & \textbf{0.69 ± 0.14} & 0.67 ± 0.13 & 0.61 ± 0.15 \\
TS+LR      & 0.62 ± 0.15  & 0.67 ± 0.15 & \textbf{0.68 ± 0.13} & 0.62 ± 0.15 \\
TS+SVM     & 0.61 ± 0.16 & \textbf{0.66 ± 0.14} & \textbf{0.66 ± 0.14} & 0.61 ± 0.15 \\
\hline
\multicolumn{5}{l}{\footnotesize \textbf{Lee2019}}\\
ACM+TS+SVM & 0.62 ± 0.14 & \textbf{0.70 ± 0.13} & 0.67 ± 0.14 & 0.66 ± 0.12 \\
CSP+LDA    & 0.63 ± 0.15 & \textbf{0.63 ± 0.16} & 0.63 ± 0.16 & 0.63 ± 0.14 \\
CSP+SVM    & 0.62 ± 0.20 & \textbf{0.65 ± 0.16} & 0.63 ± 0.15 & 0.60 ± 0.13 \\
FBCSP      & 0.63 ± 0.16 & \textbf{0.65 ± 0.16} & 0.64 ± 0.17 & 0.60 ± 0.13 \\
TS+EL      & 0.64 ± 0.17 & \textbf{0.69 ± 0.15} & 0.68 ± 0.15 & 0.63 ± 0.11 \\
TS+LR      & 0.61 ± 0.13 & \textbf{0.70 ± 0.13} & 0.63 ± 0.13 & 0.64 ± 0.11 \\
TS+SVM     & 0.60 ± 0.14 & 0.70 ± 0.13 & \textbf{0.71 ± 0.13} & 0.64 ± 0.11 \\
\hline
\multicolumn{5}{l}{\footnotesize \textbf{ALEXMI}}\\

ACM+TS+SVM & \textbf{0.60 ± 0.11} & 0.52 ± 0.12 & 0.49 ± 0.14 & 0.53 ±0.10 \\
CSP+LDA    & 0.43 ± 0.13 & \textbf{0.44 ± 0.15} & 0.43 ± 0.15 & 0.43 ± 0.15 \\
CSP+SVM    & \textbf{0.65 ± 0.20} & 0.49 ± 0.12 & 0.43 ± 0.13 & 0.45 ± 0.12 \\
FBCSP      & 0.53 ± 0.10 & \textbf{0.55 ± 0.12} & 0.47 ± 0.13 & 0.50 ± 0.11 \\
TS+EL      & 0.40 ± 0.12 & \textbf{0.43 ± 0.10} & 0.38 ± 0.13 & 0.35 ± 0.06 \\
TS+LR      & \textbf{0.51 ± 0.10} & 0.50 ± 0.08 & 0.46 ± 0.14 & 0.48 ± 0.11 \\
TS+SVM     & \textbf{0.50 ± 0.13} & 0.47 ± 0.12 & 0.47 ± 0.13 & 0.42 ± 0.11 \\
\hline
\multicolumn{5}{l}{\footnotesize *Window size for BCIC 2a, Lee2019, and AlexMI was 288,}\\
\multicolumn{5}{l}{\footnotesize
100, and 30, respectively.}
\end{tabular}%
}
\end{minipage}%
\hfill%
\begin{minipage}[t]{0.4\textwidth}
\centering

\captionof{table}{Overall Accuracy (Schirrmeister 2017)}
\label{tab:Schirrmeister}
\resizebox{\linewidth}{!}{%
\begin{tabular}{ l c c }
\hline
Models & Offline & Window (100) \\
\hline
ACM+TS+SVM & \textbf{0.79 ± 0.10}  & 0.61 ± 0.10  \\
CSP+LDA    & \textbf{0.61 ± 0.10}  & 0.58 ± 0.09  \\
CSP+SVM    & \textbf{0.61 ± 0.09}  & 0.57 ± 0.08  \\
FBCSP      & \textbf{0.70 ± 0.12}  & 0.54 ± 0.09  \\
TS+EL      & \textbf{0.73 ± 0.13}  & 0.63 ± 0.11  \\
TS+LR      & \textbf{0.69 ± 0.10} & 0.62 ± 0.10 \\
TS+SVM     & \textbf{0.70 ± 0.12}  & 0.60 ± 0.11 \\
\hline
\end{tabular}%
}
\vspace{2em}

\captionof{table}{BCIC Window Comparison}
\label{tab:bcic_window}
\vspace{0.5em}

\footnotesize
\setlength{\tabcolsep}{3pt}
\renewcommand{\arraystretch}{1}

\resizebox{\linewidth}{!}{%
\begin{tabular}{ l c c c }
\hline
Models & W(200) & W(288) & W(350) \\
\hline
ACM+TS+SVM & 0.74 ± 0.13 & 0.76 ± 0.13 & \textbf{0.76 ± 0.11} \\
CSP+LDA    & \textbf{0.60 ± 0.13} & 0.59 ± 0.13 & \textbf{0.60 ± 0.13} \\
CSP+SVM    & \textbf{0.60 ± 0.14} & \textbf{0.60 ± 0.14} & \textbf{0.60 ± 0.14} \\
FBCSP      & 0.68 ± 0.16 & \textbf{0.70 ± 0.15} & \textbf{0.70 ± 0.15} \\
TS+EL      & 0.64 ± 0.14 & 0.67 ± 0.13 & \textbf{0.68 ± 0.14} \\
TS+LR      & 0.67 ± 0.15 & \textbf{0.68 ± 0.13} & \textbf{0.68 ± 0.13} \\
TS+SVM     & 0.65 ± 0.14 & \textbf{0.66 ± 0.14} & \textbf{0.66 ± 0.14} \\
\hline
\end{tabular}%
}

\end{minipage}

\end{table}

Table~\ref{tab:bcic} reports overall accuracy on BCIC 2a for offline training versus time-ordered adaptation (average across all subjects). All pipelines improve under the incremental regime, which is consistent with temporal non-stationarity in this dataset. ACM+TS+SVM is best overall (0.69 offline to 0.77 incremental). TS-based models also gain substantially (about 0.59 offline to 0.67 with rolling window to 0.69 incremental), while CSP baselines remain lower even after adaptation (about 0.54 offline to 0.59 with rolling window to 0.61 incremental). FBCSP performed as the second-best model in this dataset.

Lee2019 shows a broadly similar pattern. Online approaches outperformed offline approaches. However, the models exhibited a different ranking order. ACM+TS+SVM is among the best online models, but it is not the best offline model. FBCSP remained competitive in the offline setting (0.63, close to the best offline model, TS+EL, at 0.64), yet it was outperformed by other models in the incremental and cold-start settings. TS-based models performed generally better in this dataset. Aggregating across the two largest streams (BCIC 2a and Lee2019, seven pipelines each), the best online regime improves 13 of the 14 model--dataset pairs relative to the frozen offline baseline; only ALEXMI reverses this trend (see below).

ALEXMI, however, deviates from the previously described behaviour. Online settings generally underperformed offline settings. Here, the best adaptive models are still ACM+TS+SVM and FBCSP. Because ALEXMI provides only 20 trials per class, the recent-window and cold-start regimes quickly run into data starvation, which we revisit through the data-valuation analysis.

Table~\ref{tab:Schirrmeister} shows the performance of the models on Schirrmeister2017, the dataset with the highest number of channels (128, about twice as many as Lee2019; we use a 32-channel subset, see Datasets). Because of the computational cost and the unequal number of trials per subject, only the frozen model and a 100-trial rolling window were tested on this dataset. The window is lower for all seven pipelines. Because expanding-window and cold-start results were not computed, this table supports only the conclusion that this particular 100-trial memory budget is too small under the tested protocol.

Table~\ref{tab:bcic_window} shows a comparison between multiple window sizes using the BCIC 2a dataset. Accuracy changes modestly between 200 and 350 trials, and the maximizing window differs by model. This sensitivity suggests that the optimal amount of forgetting is model-dependent rather than universal. On BCIC~2a and Lee2019, cold-start accuracy never exceeds the incremental regime (with one tie), quantifying the cost of minimal initial calibration.
Nevertheless, the cold-start regime shows that some pipelines can reach nontrivial performance after only a small initial labeled prefix, followed by label-revealed online updates.

\subsection*{Prequential dynamics: frozen vs.\ online refit}

\begin{figure}[htbp]
  \centering
  \includegraphics[width=0.92\linewidth]{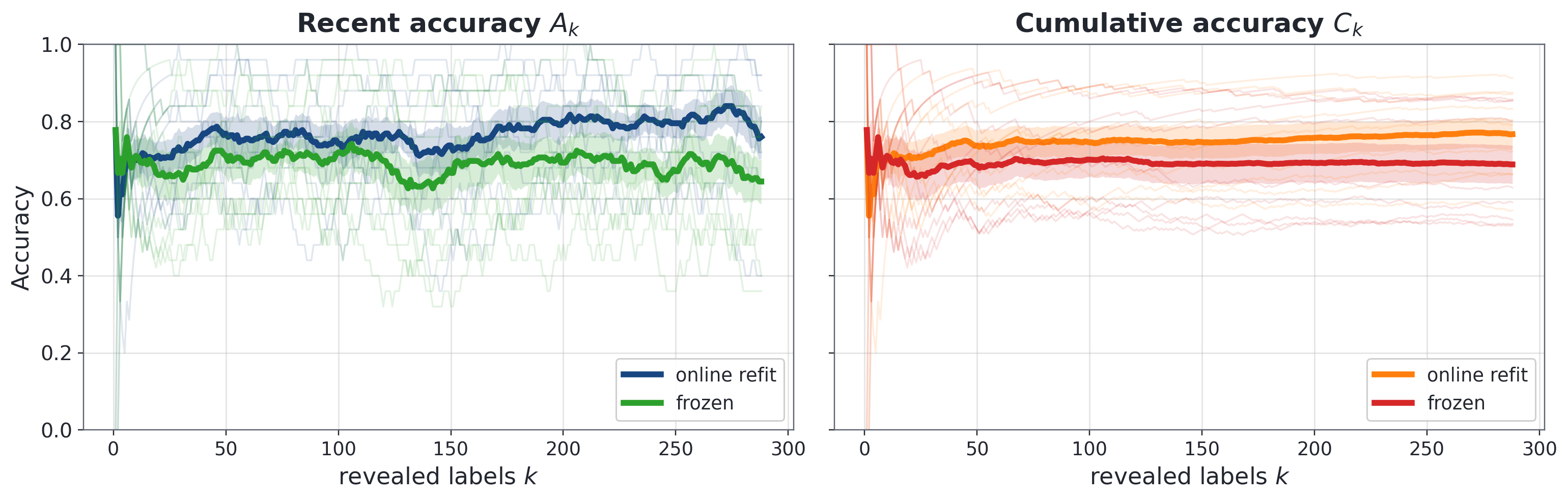}
  \caption{Prequential accuracy on BCIC 2a for ACM+TS+SVM, frozen vs.\ online refit (9 subjects; bold curves are the across-subject mean, shaded bands the standard error, thin curves individual subjects). Left: trailing-window accuracy $A_k$ (Eq.~\ref{eq:ak}, $w=25$). Right: cumulative accuracy $C_k$ (Eq.~\ref{eq:ck}). Online refit remains above frozen for most of the stream and the advantage grows over time.}
  \label{fig:acm_curves}
\end{figure}

To see when the benefit of updating appears, we track the prequential curves of two representative pipelines on BCIC 2a (9 subjects, 288 evaluated online steps), contrasting a \emph{frozen} model (trained once on the initial prefix, never updated during the stream) with \emph{online refit} (retrained after every revealed label, i.e.\ the expanding-window scheme). Figure~\ref{fig:acm_curves} shows the covariance-based ACM+TS+SVM: online refit stays above frozen for most of the stream: the cumulative score $C_k$ is higher at 277 of 288 steps (mean stepwise gain $+0.055$), and the gap widens as more labels arrive.

Table~\ref{tab:refit} reports the final scores for both pipelines. For ACM+TS+SVM, online refit improves final \(C_k\) by 0.078 and final \(A_k\) by 0.116. After Holm correction across the four paired Wilcoxon tests, the \(A_k\) difference remains significant (\(p_{\mathrm{Holm}}=0.046\)), while the \(C_k\) difference does not (\(p_{\mathrm{Holm}}=0.059\)). FBCSP shows smaller gains (\(+0.034\) for \(C_k\), \(+0.022\) for \(A_k\)), neither significant after correction (\(p_{\mathrm{Holm}}=0.258\) and \(0.499\), respectively). Both pipelines show positive mean changes under online updating, but the gain is larger for the covariance-based ACM+TS+SVM than for FBCSP. 

\begin{table}[htbp]
\centering
\caption{Final prequential accuracy on BCIC 2a (9 subjects, 288-step stream), frozen vs. online refit. Holm-adjusted paired Wilcoxon tests give \(p=0.059\) and \(0.046\) for ACM+TS+SVM final \(C_k\) and \(A_k\), and \(p=0.258\) and \(0.499\) for FBCSP.}
\label{tab:refit}
\small
\begin{tabular}{lcccc}
\toprule
 & \multicolumn{2}{c}{ACM+TS+SVM} & \multicolumn{2}{c}{FBCSP}\\
\cmidrule(lr){2-3}\cmidrule(lr){4-5}
Metric & Frozen & Online refit & Frozen & Online refit\\
\midrule
Final $C_k$ & 0.689 & \textbf{0.767} & 0.639 & \textbf{0.673}\\
Final $A_k$ & 0.644 & \textbf{0.760} & 0.627 & \textbf{0.649}\\
\bottomrule
\end{tabular}
\end{table}

\FloatBarrier
\paragraph{Analysis}

Based on the results reported in Tables~\ref{tab:bcic} and~\ref{tab:bcic_window}, we further investigated how the importance of individual training samples changes with respect to performance on future test data. We focus on the BCIC 2a dataset. Comparing the offline setting to the sliding-window setting (window size 288) on BCIC 2a, we observe that, even when the training set sizes are matched, models trained on more recent data achieve better performance on subsequent datapoints. This suggests a non-stationary setting where recency of training data matters.

To examine this effect in more detail, we performed a Shapley-based contribution analysis. Shapley values were estimated using the Truncated Monte Carlo Shapley method \cite{ghorbani2019data}, as implemented in the pyDVL library \cite{TransferLab_team_pyDVL_2024}. We split the timeline into two segments: the first 432 samples (288 + 144) served as the training pool, and the final 144 samples served as the test set. The training pool was further partitioned into three consecutive blocks of equal size (144 samples each), denoted as part 1, part 2, and part 3, where part 3 is temporally closest to the test segment. We then quantified how samples from each training block contributed to the test performance.

For each subject and seed, we computed the total Shapley contribution of each training block by summing the 144 per-sample Shapley values within that block. Random seeds are repeated computations, not independent participants. We therefore averaged the three seed-specific block sums within each subject and conducted paired, two-sided t-tests across the nine subject means. We considered part 1 vs. part 3, part 2 vs. part 3, and part 1 vs. part 2, and adjusted the three p-values using the Holm–Bonferroni procedure.

Consistent with the recency hypothesis, the training block closest to the test segment exhibited the largest total Shapley value, indicating higher marginal contribution to the test performance. Fig.~\ref{fig:shapley_subtable} summarizes the paired comparisons between the summed Shapley values of the three training blocks. Both comparisons against part 3 are statistically significant after Holm correction (part 1 vs.\ part 3: $p_{\mathrm{Holm}}=0.0176$; part 2 vs.\ part 3: $p_{\mathrm{Holm}}=0.0176$), whereas the comparison between part 1 and part 2 is not significant ($p_{\mathrm{Holm}}=0.7507$).

Finally, we analyzed sample-level influence as a function of temporal distance. For each training point, we related its leave-one-out influence on the test trials to its temporal distance from them and averaged the resulting values within distance bins. Figure~\ref{fig:shapley_subplot} shows the average influence profile as a function of distance. To reduce noise due to limited support, we omit the first and last 10 distance bins, which contain only a small number of train–test pairs.

\newlength{\panelh}
\setlength{\panelh}{4.6cm}

\begin{figure}[htbp]
  \centering

  \begin{subfigure}[b]{0.54\textwidth}
    \centering

    \begin{minipage}[c][\panelh][c]{\linewidth}
        \centering
        \scriptsize
        \setlength{\tabcolsep}{2.5pt}

        \begin{tabular}{@{}lrrrrrr@{}}
            \toprule
            Comparison
            & $n$
            & mean(L)
            & mean(R)
            & $d_z$
            & $p$
            & $p_{\mathrm{Holm}}$ \\
            \midrule
            \textbf{part-1 vs part-3}
            & $9$ & $0.191$ & $0.302$ & $-1.225$ & $0.0063$ & $0.0176$ \\

            \textbf{part-2 vs part-3}
            & $9$ & $0.184$ & $0.302$ & $-1.240$ & $0.0059$ & $0.0176$ \\

            \textbf{part-1 vs part-2}
            & $9$ & $0.191$ & $0.184$ & $0.110$ & $0.7507$ & $0.7507$ \\
            \bottomrule
        \end{tabular}
    \end{minipage}

    \caption{
        Subject-level paired comparisons of block-summed sample Shapley
        values after averaging the three seeds within each subject.
        Here, mean(L) and mean(R) are the across-subject means and
        $d_z$ is paired-sample Cohen's effect size.
    }
    \label{fig:shapley_subtable}
\end{subfigure}
  \hfill
  \begin{subfigure}[b]{0.43\textwidth}
    \centering
    \includegraphics[height=\panelh,width=\linewidth,keepaspectratio]{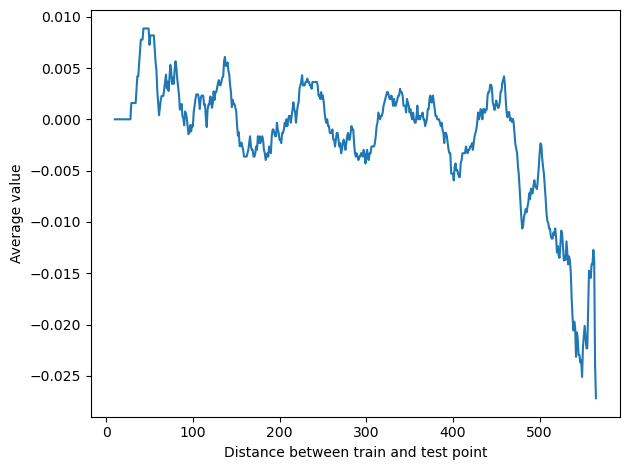}
    \caption{Influence vs.\ temporal distance.}
    \label{fig:shapley_subplot}
  \end{subfigure}

  \caption{Training-sample contribution analysis on BCIC. Left: paired tests comparing summed Shapley values across training blocks. Right: influence profile as a function of temporal distance.}
  \label{fig:shapley_table_plot}
\end{figure}

\textbf{Exact block-level valuation across datasets.}
The Monte-Carlo estimate above values individual samples grouped into three coarse blocks. To value \emph{whole temporal blocks} directly and to obtain an \emph{exact} rather than sampled Shapley value, we instead treat each contiguous block of the ordered training pool as a single player. With only $P$ blocks ($P\in\{5,6\}$), the $2^{P}$ coalitions can be enumerated exhaustively, so each block's Shapley value is computed without Monte-Carlo error. This set of experiments was done for ALEXMI, BCIC 2a and Schirrmeister datasets, using both the Riemannian-based (ACM+TS+SVM) and spatial-filtering (FBCSP) pipelines and an analogous test-suffix protocol as above.

\begin{figure}[htbp]
  \centering
  \includegraphics[width=0.72\linewidth]{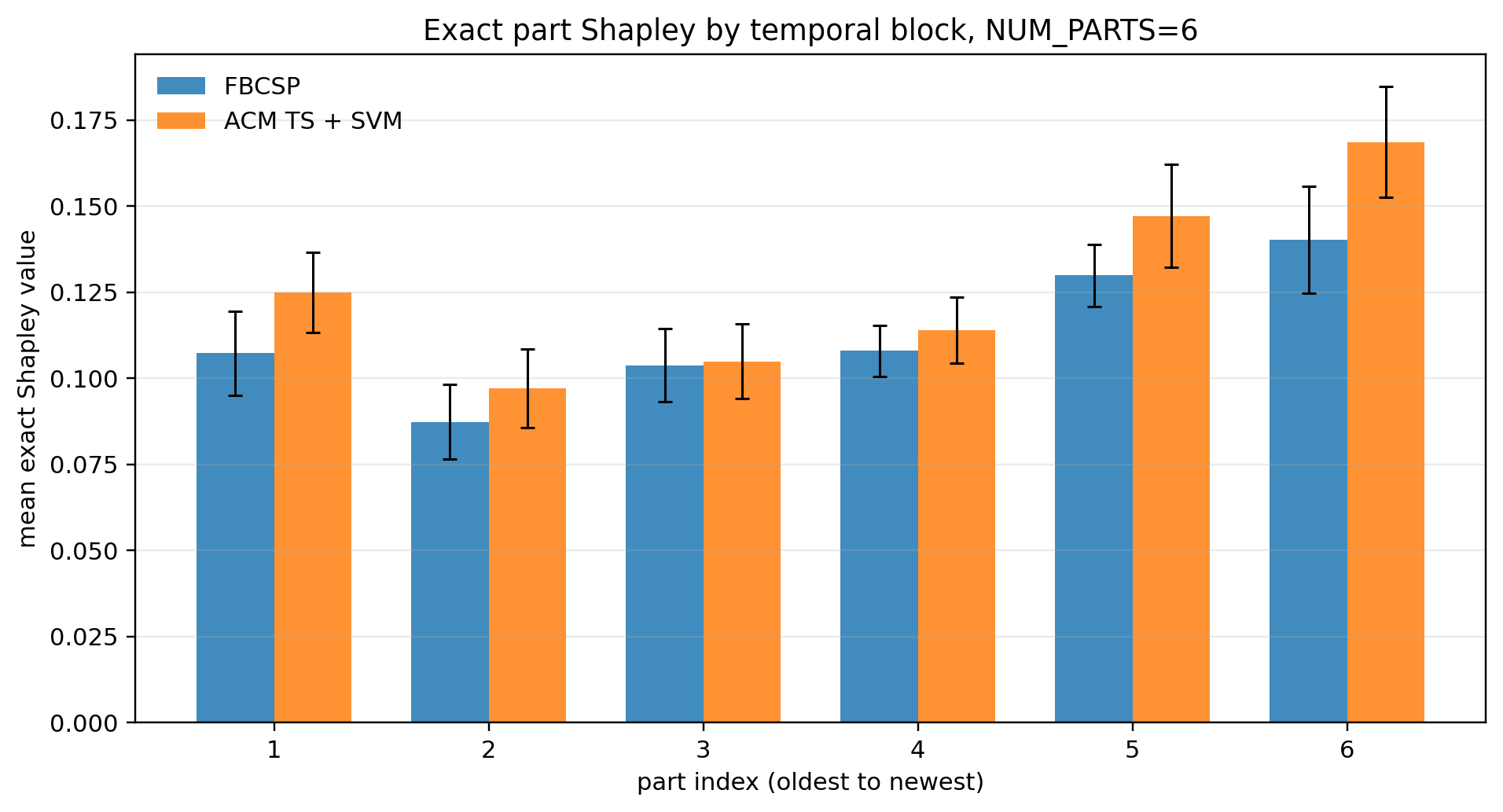}
  \caption{Exact block-level (part) Shapley value on BCIC 2a. The 432-trial training pool is split into $P=6$ contiguous temporal blocks (oldest to newest), each treated as a single Shapley player, so the value is computed by exact enumeration of all $2^{6}$ coalitions. Bars are the across-run mean $\pm$ standard error for FBCSP and ACM+TS+SVM. The newest blocks receive the largest credit in both pipelines.}
  \label{fig:exact_bcic}
\end{figure}

\begin{figure}[htbp]
  \centering
  \begin{subfigure}[b]{0.48\textwidth}
    \centering
    \includegraphics[width=\linewidth]{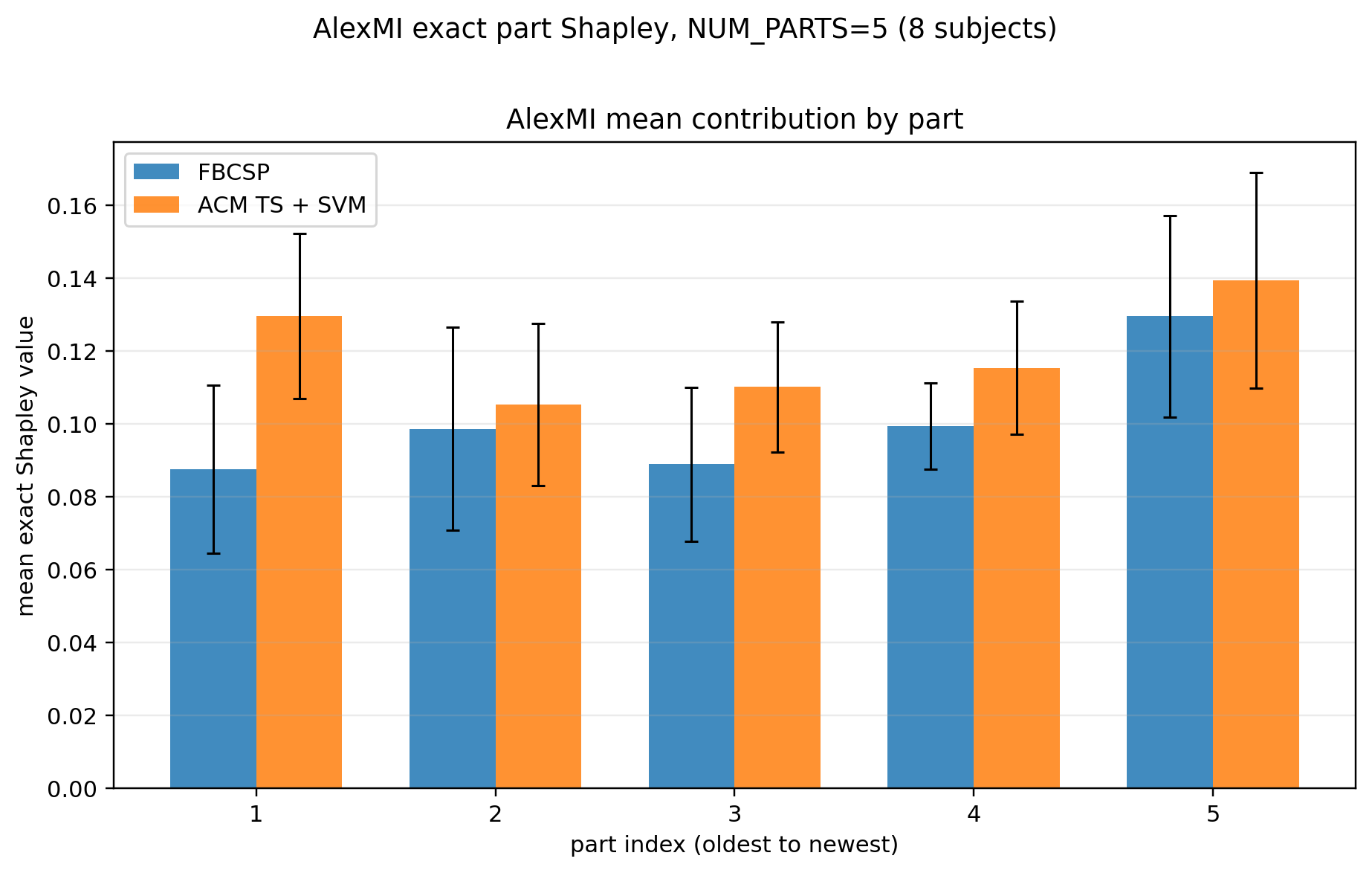}
    \caption{ALEXMI (8 subjects, $P=5$, block size 6).}
    \label{fig:exact_alexmi}
  \end{subfigure}
  \hfill
  \begin{subfigure}[b]{0.48\textwidth}
    \centering
    \includegraphics[width=\linewidth]{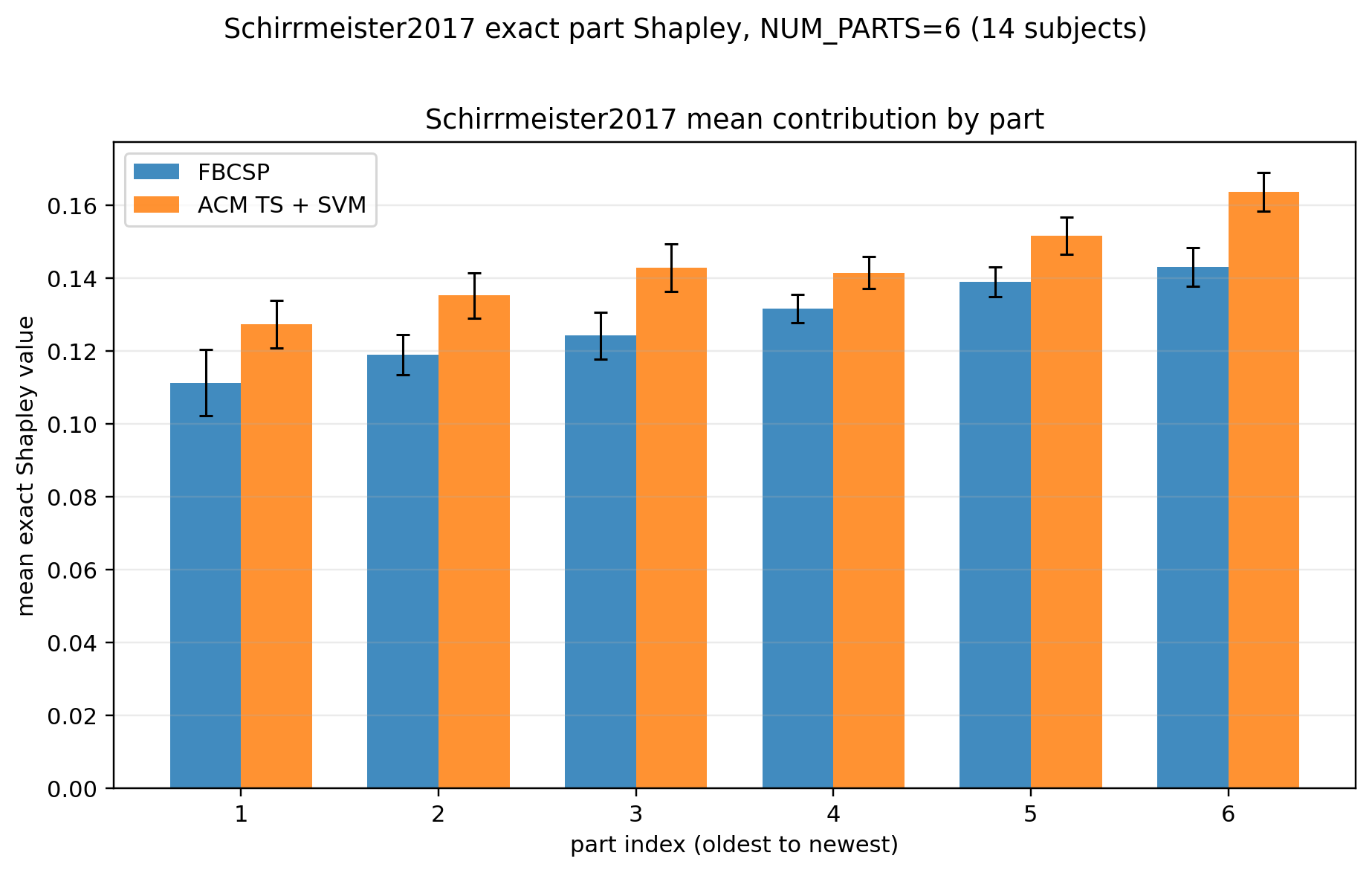}
    \caption{Schirrmeister 2017 (14 subjects, $P=6$, mean block size $\approx120$).}
    \label{fig:exact_schir}
  \end{subfigure}
  \caption{Cross-dataset comparison of the exact block-level Shapley valuation. The newest temporal block carries the largest mean value in both pipelines and on both datasets; the gradient is flatter on the small ALEXMI stream and sharper on the larger Schirrmeister 2017 streams.}
  \label{fig:exact_cross}
\end{figure}

Figure~\ref{fig:exact_bcic} shows BCIC 2a with the same 432-trial pool now split into $P=6$ blocks of 72 trials. In both pipelines, the two newest blocks receive more credit than the rest, an effect that is most pronounced for ACM+TS+SVM. A similar pattern is observed on the other datasets (Fig.~\ref{fig:exact_cross}). On the small ALEXMI stream (8 subjects, $P=5$, blocks of 6 trials; full-pool accuracy $0.60\pm0.09$ for ACM+TS+SVM and $0.50\pm0.11$ for FBCSP), the newest block still carries the largest mean Shapley value (0.139 and 0.130 respectively), though the profile is flatter, possibly because each block holds only six trials. In the larger Schirrmeister\,2017 streams (14 subjects, $P=6$, a 32-channel subsampled montage, mean block size $\approx120$ trials; full-pool accuracy $0.86\pm0.08$ and $0.77\pm0.09$), the recency gradient is cleaner and the newest-block value is again the largest (0.164 and 0.143). Thus, the newest block receives the largest mean value in all three datasets, and the profile is sharper when blocks contain more trials; with only three datasets and no cross-dataset test, the apparent difference in profile sharpness cannot be attributed uniquely to block size.

\begin{figure}[htbp]
  \centering
  \includegraphics[width=0.62\linewidth]{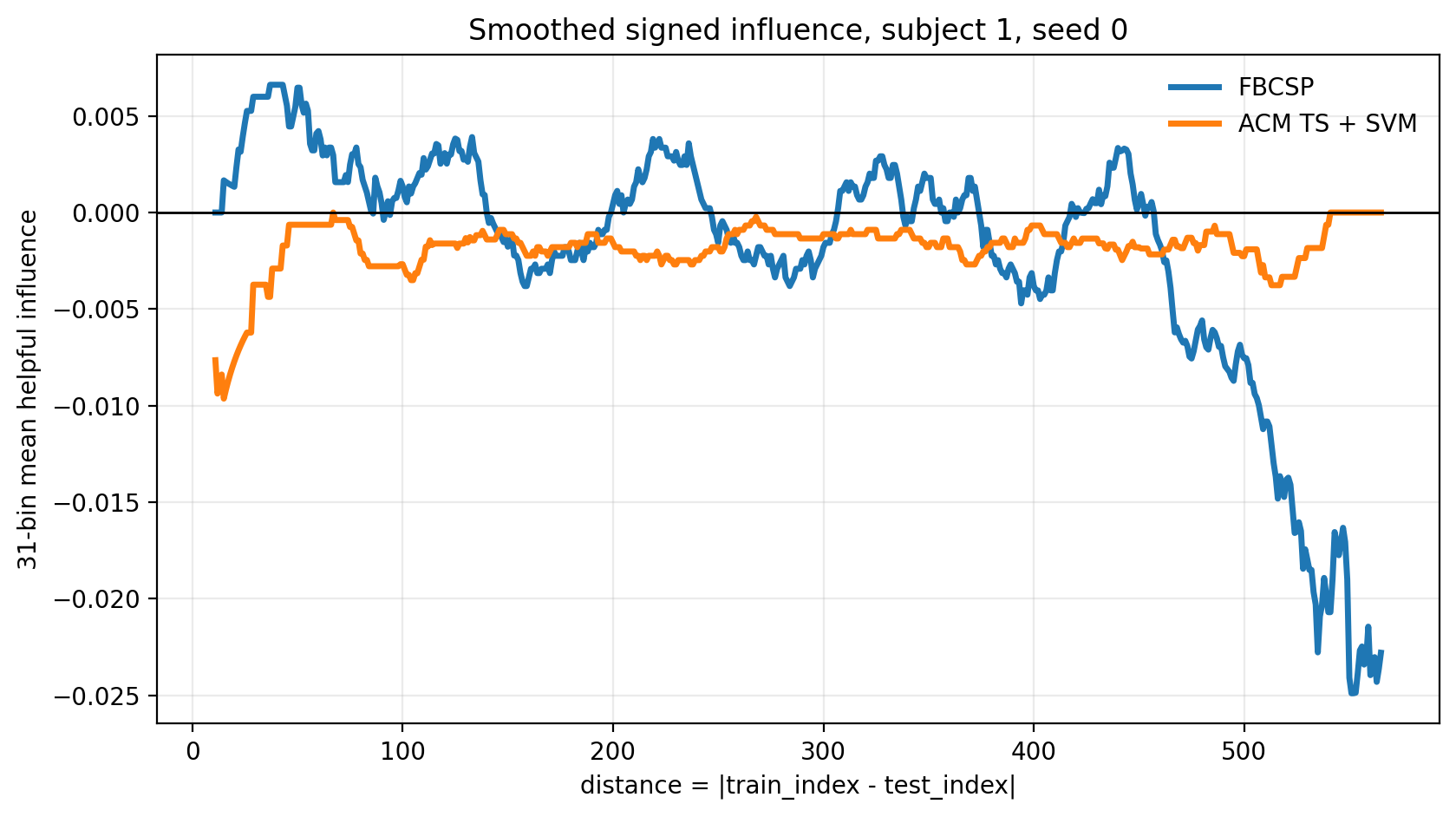}
  \caption{Smoothed signed helpful influence versus train--test temporal distance for a single subject (subject 1, seed 0). Values above zero are helpful, below zero harmful. FBCSP is slightly positive at short distances and turns negative in the far tail, whereas ACM+TS+SVM stays much closer to zero across the axis, indicating that FBCSP is more sensitive to temporal distance for this subject.}
  \label{fig:far_tail}
\end{figure}

\textbf{Local influence, block value, and forgetting.}
Older blocks retain \emph{positive} Shapley value, which helps reconcile two analyses that initially appear to conflict. The per-sample influence-vs-distance curve (Fig.~\ref{fig:shapley_subplot}) stays close to zero over most of the distance range and turns negative only in the far tail, and for FBCSP some individual old samples even show a mildly negative marginal contribution at large temporal distance on single subjects (Fig.~\ref{fig:far_tail}). Yet the rolling-window results show that discarding old data can \emph{hurt} FBCSP: the rolling window falls below the expanding-history incremental scheme on Lee2019 and ALEXMI (Table~\ref{tab:bcic}), and the 100-trial window is 16 points below the frozen model on Schirrmeister~2017 (Table~\ref{tab:Schirrmeister}), although on BCIC~2a the 288-trial window is slightly better (0.70 vs.\ 0.69). These observations are not contradictory: leave-one-out influence measures the local marginal effect of removing a single point (and, being aggregated over temporal distance, does not preserve the block-level coalition structure), whereas switching to a rolling window changes the entire training policy, including sample count, class balance, and the pooled covariance/spatial-filter estimates on which FBCSP especially depends. This suggests forgetting selectively rather than discarding all old data: recent trials appear to be the most valuable, but enough history should be retained for stable estimation.

\FloatBarrier
\section*{Conclusion}

EEG-based brain–computer interfaces (BCIs) face strong non-stationarity, causing static decoding models to degrade over time. We frame online BCI decoding as a label-revealed continual-learning problem and introduce a lightweight benchmark based on prequential evaluation. Across multiple EEG datasets, we compare CSP- and covariance-based pipelines under streaming updates and analyze adaptation dynamics. In the presented experiments, online refit can give up to about 18\% relative accuracy improvement over a frozen offline model.

\textbf{Online adaptation helps on the larger datasets but not on the smallest.} BCIC 2a is the largest MI dataset (in terms of number of datapoints) used in the set of experiments. It has 144 trials per class and four classes. Lee2019 is the second largest dataset, having 100 trials per class per session and two classes. ALEXMI is the smallest dataset by far, with 20 trials per class and three classes. Across the experiments, BCIC 2a and Lee2019 show broadly similar behavior: online adaptation usually improves performance relative to the frozen offline baseline, although the best adaptive regime depends on the model. ALEXMI deviates from this pattern. On this small dataset, rolling-window and cold-start regimes generally underperform the offline baseline, and the incremental regime underperforms offline for 4 of the 7 pipelines. This suggests that the number of available trials is an important factor when choosing an online update policy.

\textbf{Retaining history is often useful, but the optimal amount of forgetting is model-dependent.} The expanding-window regime retains all previously observed labeled trials, whereas the rolling-window regime explicitly forgets older trials as new labels arrive. Across the larger streams, retaining history is often beneficial, plausibly because it preserves sample count, class balance, and stable parameter estimation. However, rolling windows can still match or outperform expanding windows for some model--dataset pairs, indicating that older data may become less useful under drift. The update policy should therefore be treated as a tunable part of the decoding pipeline rather than as a universal choice.

\textbf{Recent trials carry the most value, but old trials are not disposable.} A Shapley-based valuation over temporal blocks assigns the largest mean credit to the most recent block on every dataset included in the Shapley analysis. This supports the interpretation that temporal non-stationarity contributes to the online gains. At the same time, older blocks retain positive aggregate value, possibly because they stabilize the fitted representation (tangent-space reference estimates, classifier parameters, and the pooled covariance estimates used for CSP/FBCSP spatial filters). This points to selective forgetting rather than uniform windowing: recent labels appear most valuable, while enough history should be retained for stable estimation.

Future work will explore lightweight deep neural networks in a continual-learning setting, cross-subject and transfer-based initialization to reduce calibration, and extensions to additional BCI paradigms such as P300 and SSVEP.

{\small
\bibliographystyle{IEEEtran}
\bibliography{bibliography}
}

\end{document}